\pdfoutput=1
\documentclass[11pt]{article}

\usepackage[preprint]{acl}

\usepackage{times}
\usepackage{latexsym}

\usepackage[T1]{fontenc}
\usepackage[utf8]{inputenc}

\usepackage{microtype}

\usepackage{inconsolata}

\usepackage{graphicx}

\usepackage{booktabs}       % professional-quality tables
\usepackage{amsfonts}       % blackboard math symbols
\usepackage{natbib}
\usepackage{doi}

\usepackage{multirow}
\usepackage{threeparttable}
\usepackage{subcaption} 
\usepackage{array} 
\usepackage{amsmath}

\usepackage{amsthm}
\usepackage{float} 
\usepackage{autobreak}
\usepackage[normalem]{ulem}
\useunder{\uline}{\ul}{}

\title{Towards Reward Fairness in RLHF: From a Resource Allocation Perspective}

\author{
 \textbf{Sheng Ouyang\textsuperscript{1,2,3,4}}\thanks{~~Work done during an internship at Kuaishou Technology.},
 \textbf{Yulan Hu\textsuperscript{4}}$^\dagger$,
 \textbf{Ge Chen\textsuperscript{4,5$*$}},
 \textbf{Qingyang Li\textsuperscript{4}},
 \textbf{Fuzheng Zhang\textsuperscript{4}},
 \textbf{Yong Liu\textsuperscript{1,2,3,4}}\thanks{~~Corresponding authors.}
\\
 \textsuperscript{1}Gaoling School of Artificial Intelligence, Renmin University of China \\
 \textsuperscript{2}Beijing Key Laboratory of Research on Large Models and Intelligent Governance \\
 \textsuperscript{3}Engineering Research Center of Next-Generation Intelligent Search and Recommendation, MOE \\
 \textsuperscript{4}Kuaishou Technology \\
 \textsuperscript{5}University of Chinese Academy of Sciences
\\
 \small
 {
   \{ouyangsheng, liuyonggsai\}@ruc.edu.cn, \{huyulan, liqingyang, zhangfuzheng\}@kuaishou.com, chenge221@mails.ucas.ac.cn
 }
}

\begin{document}
\maketitle
\begin{abstract}

Rewards serve as proxies for human preferences and play a crucial role in Reinforcement Learning from Human Feedback (RLHF). However, if these rewards are inherently imperfect, exhibiting various biases, they can adversely affect the alignment of large language models (LLMs). In this paper, we collectively define the various biases present in rewards as the problem of reward unfairness. We propose a bias-agnostic method to address the issue of reward fairness from a resource allocation perspective, without specifically designing for each type of bias, yet effectively mitigating them. Specifically, we model preference learning as a resource allocation problem, treating rewards as resources to be allocated while considering the trade-off between utility and fairness in their distribution. We propose two methods, Fairness Regularization and Fairness Coefficient, to achieve fairness in rewards. We apply our methods in both verification and reinforcement learning scenarios to obtain a fairness reward model and a policy model, respectively. Experiments conducted in these scenarios demonstrate that our approach aligns LLMs with human preferences in a more fair manner. Our data and code are available at
\url{https://github.com/shoyua/Towards-Reward-Fairness}.
\end{abstract}

\section{Introduction} \label{sec:intro}

RLHF~\cite{ouyang2022training,rlhfsurvey,rlhfflow} has significantly advanced the alignment of  LLM outputs with human preferences, ensuring that the responses are helpful, harmless, and honest~\cite{hh-rlhf,Trustworthiness}. The reward model (RM)~\cite{stiennon2020learning,ouyang2022training,robust_rm} plays a crucial role in this process by providing a quantitative metric that measures the degree to which the model outputs align with human preferences. This metric guides the LLM in producing outputs that are more consistent with human preference.
\begin{figure}
\centering
    % \begin{minipage}[c]{0.47\textwidth}
    % \centering
    \includegraphics[width=\linewidth]{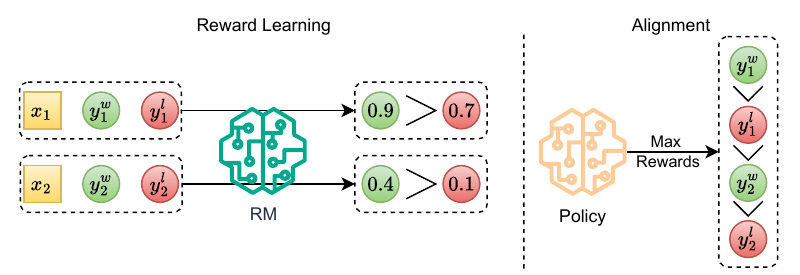}
    \caption{Rewards unfair problem in RLHF.}
    \label{fig:introduction}
\end{figure}

One of the key reasons for the success of RLHF is the assumption that the reward model can accurately represent and measure actual preferences~\cite{kim2024rethinking}. However, if the reward model itself is biased~\cite{bias1,bias2,reber2024rate}, it can lead to policy models exhibiting behaviors that do not align with human preferences~\cite{scalerm,2024alpacafarm}. 
Consider Figure~\ref{fig:introduction}, where we denote a preference pair data as $(x, y^w, y^l)$, representing a query, a preferred and dispreferred response, respectively. When training a reward model following the Bradley-Terry model~\cite{BT_model} with such preference data, the output appears reasonable because for each $x$, the reward of $y^w$ is greater than that of $y^l$. However, using this reward model to guide the training of a policy model can be problematic. The policy model aims to maximize rewards, and in this scenario, the reward for $y_1^l$ is greater than the reward for $y_2^w$, even though $y_1^l$ is dispreferred and $y_2^w$ is preferred. 
From a general perspective, if \((x_1, y_1^w, y_1^l)\) and \((x_2, y_2^w, y_2^l)\) come from different data types, such bias in their reward distribution  can steer the model to favor one type of data over another.
We define this issue as ``\textbf{reward unfairness}''.

We interpret various reward biases from the perspective of reward unfairness, including length bias~\cite{acl23length,disentangling}, category bias~\cite{padmakumar2024beyond}, and social bias~\cite{li2023survey}. For instance, when the rewards distribution varies significantly across data of different lengths or categories, resulting in reward unfairness, it manifests as length bias or category bias respectively. Existing work~\cite{disentangling,chenodin} has addressed these biases by proposing targeted methods to mitigate them.
\citet{disentangling,chenodin,yangrewards,padmakumar2024beyond} have employed techniques such as length regularization  to mitigate the effects of length bias. These methods adjust the reward distribution to prevent models from favoring longer responses, thereby ensuring more fair outputs.
On the other hand, category bias has not been as widely acknowledged. However, some studies have implicitly addressed this issue. For instance, the work on learning diverse preferences~\cite{yangrewards,padmakumar2024beyond} and model ensemble \cite{warm,warp} indirectly reduce the impact of category bias. These work promote a more varied and representative set of outputs, which helps minimize the skewness introduced by category-specific biases.

However, these works are specifically designed to address particular biases and lack the ability to transfer solutions across different types of biases. In this paper, we propose a unified perspective that considers these biases as manifestations of a broader issue: reward unfairness. To address this comprehensively, we introduce the reward fairness framework.
Firstly, we model preference learning as a resource allocation problem \cite{katoh1998resource}. In this framework, we define the rewards in preference learning as the resources to be allocated. The extent to which these rewards reflect human preferences is defined as utility, while the consistent distribution of rewards across the data is defined as fairness. We employ a unified fairness function to measure the fairness of the rewards distribution. This approach seeks to achieve a trade-off between fairness and utility. We propose two methods to obtain fairness rewards: Fairness Regularization and Fairness Coefficient.
We then apply these methods in two scenarios: Fairness Rewards for Verification and Fairness Rewards for Reinforcement Learning (RL). 
We conclude our contributions as following:
\paragraph{Unified Perspective from Reward Unfairness} 
We introduce a novel perspective that frames various biases as specific instances of the broader problem of reward unfairness. This unified view fosters a more comprehensive understanding and approach to addressing these biases.

\paragraph{Reward Fairness Framework} We propose the reward fairness framework from a resource allocation perspective to systematically address reward unfairness, aiming to balance fairness and utility in reward distribution.

\paragraph{Application to Verification and RL} We apply our proposed methods in two scenarios: (a) Fairness Rewards for Verification, which focuses on training a fairness RM, and (b) Fairness Rewards for RL, which aims to train a policy model that implicitly incorporates fair rewards.  Our fairness rewards methods can be seamlessly integrated with existing RM and RL methods.

\section{Related Work}

RLHF has become the standard approach for aligning LLMs with human preferences. RLHF can be decomposed into two main components: Reward Learning and RL Finetune.
\subsection{Reward Learning}
The reward model is a crucial component of RLHF, providing a quantitative metric to guide alignment with human preferences. 
Reward models typically follow the Bradley-Terry model \cite{BT_model}, but there are also approaches based on regression paradigms \cite{wang2024interpretable,wang2024helpsteer2}  and the ``LLM as a judge'' approach \cite{zhang2024generative,zheng2023judging}. However, \citet{hou2021imperfect,kim2024rethinking,reber2024rate} have identified that reward models are imperfect proxies for human preferences, exhibiting various issues such as length bias \cite{acl23length} and reward hacking \cite{skalse2022defining}. \citet{acl23length,chenodin} have found that the results of reward models are influenced by the length of the input, and they have attempted to decouple this relationship during training to mitigate its effects. 
Fast RL~\cite{li2024optimizing} is closest to our method, however Fast RL is an ensemble method that considers fairness between different reward functions.
\subsection{RL Finetune}
RL Finetuning \cite{rlhfflow} generally involves using reinforcement learning techniques, guided by the reward model, to train the policy model. Algorithms such as proximal policy optimization (PPO) \cite{ppo} and group relative policy optimization (GRPO) \cite{grpo} are commonly used. 
There is also a category of work that omits the reward model and directly learns preference, such as direct preference optimization (DPO) \cite{dpo}, Kahneman-Tversky optimization (KTO)~\cite{KTO}, and SimPO \cite{Simpo}. These methods are more efficient and stable compared to PPO-based approaches. Although they do not involve the reward model in training, they implicitly fit rewards to align with human preferences. \citet{lu2024eliminating,liu2024length,dubois2024length}  have observed that aligned models tend to generate longer responses, which introduces a length bias. To mitigate this issue, they have proposed methods such as length regularization \cite{disentangling}.

\section{Preliminaries}

\paragraph{Reward Model}
In RLHF, RM acts as a proxy for human preferences to rate the quality of the model output. Generally, the RM follows the  Bradley-Terry Model \cite{BT_model} and can be formulated as:
\begin{equation}
\begin{split}
&p\left(y_{w} \succ y_{l} \mid x\right)= \\ & \frac{\exp \left(r_{\phi}\left(x, y_{w}\right)\right)}{\exp \left(r_{\phi}\left(x, y_{w}\right)\right)+\exp \left(r_{\phi}\left(x, y_{l}\right)\right)},
\end{split}
\end{equation}
where $(x,y_w,y_l)\sim \mathcal{D}$ represent a prompt , a preferred response and an dispreferred response from the preference dataset $\mathcal{D}$, respectively. $r_{\phi}(x, y)$ denotes a reward function with the parameters ${\phi}$, and this is subsequently denoted as $r_{\phi}(y)$ for simplicity. We can train a RM $r_\phi$ following the log-likelihood maximization as:
\begin{equation}
\max _{r_{\phi}} ~ \mathbb{E}_{\left(x, y_{w}, y_{l}\right) \sim \mathcal{D}}  \left[\log \sigma\left(r_{\phi}\left(y_{w}\right)- r_{\phi}\left(y_{l}\right)\right)\right],
\label{eq:rm_loss}
\end{equation}
where \(\sigma\) is the sigmoid function.

\paragraph{RL Finetune} During the RL phase \cite{jaques2017sequence}, the learned RM is used to provide feedback to the policy model $\pi_\theta$  with the parameters ${\theta}$. The optimization is formulated as:
\begin{equation}
\begin{split}
\max _{\pi_\theta} ~&\mathbb{E}_{x \sim \mathcal{D}, y \sim \pi_\theta(y \mid x)}\left[r_\phi(y)\right] \\ &-\beta {D}_{\mathrm{KL}}\left[\pi_\theta(y \mid x) \| \pi_{\mathrm{ref}}(y \mid x)\right],
\end{split}
\label{eq:rl_loss}
\end{equation}
where $\beta$ is a hyperparameter controlling the KL penalty and $\pi_{\mathrm{ref}}$ is the reference model.

\paragraph{DPO} DPO \cite{dpo} is a method used to directly optimize a policy based on preference data. The objective of DPO is to align the policy \(\pi_\theta\) with human preferences by maximizing the likelihood of preferred outcomes.
\begin{equation}
\begin{aligned}
\max _{\pi_\theta} ~& \mathbb{E}_{(x, y_w, y_l) \sim \mathcal{D}} \left[ \log \sigma \left( \beta \log \frac{\pi_\theta(y_w \mid x)}{\pi_{\mathrm{ref}}(y_w \mid x)} \right. \right. \\
& \left. \left. - \beta \log \frac{\pi_\theta(y_l \mid x)}{\pi_{\mathrm{ref}}(y_l \mid x)} \right) \right],
\end{aligned}
\label{eq:dpo_loss}
\end{equation}
where  \(\beta\) is a scaling factor.

\begin{figure}[htb]
    \centering
    \includegraphics[width=1\linewidth]{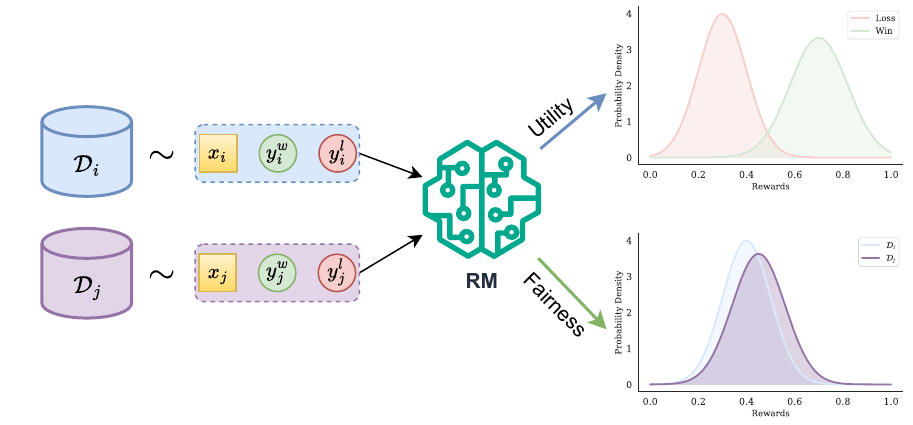}
    \caption{Objective of Fairness Rewards. \(\mathcal{D}_i\) and \(\mathcal{D}_j\) represent different data. Fairness rewards aim to obtain rewards that consider the trade-off between utility and fairness. Utility refers to the ability of the rewards to distinguish between preferred and dispreferred responses, as illustrated in the top-right figure. Fairness refers to the consistent distribution of rewards across different data, as depicted in the bottom-right figure.}
    \label{fig:overview}
\end{figure}
\section{Fairness Rewards Allocation}\label{sec:fair-reward}
In this section, we aim to obtain the fairness rewards, as shown in Figure \ref{fig:overview}. We model preference learning as a resource allocation problem that maximizes utility while ensuring the fairness of rewards.

\paragraph{Resource Allocation} Resource allocation~\cite{katoh1998resource} involves distributing resources \( R \) among entities \( i \) to optimize overall utility \( U \). The trade-off between fairness and utility is a key consideration. Utility maximization is given by:
\begin{equation}
\max  U(\mathbf{a}) \quad\quad
\text{s.t.} \quad \sum_{i} a_i \leq R,
\label{eq:utility}
\end{equation}
where $\mathbf{a}=[a_1,a_2, \dots,a_n]$ is an allocation vector and $a_i$ denotes the resources allocated to the $i$-th entity.
Fairness can be incorporated by adding a fairness constraint \( F(\mathbf{a}) \):
\begin{equation}
\begin{split}
     \max \quad & U(\mathbf{a}) \\
 \text{s.t.} \quad &\sum_{i} a_i \leq R \\
 {} &F(a) \geq  \eta
\end{split},
\label{eq:fair-utility}
\end{equation}
where \( \eta \) represents the desired level of fairness. Balancing these objectives requires careful consideration of both fairness and utility to achieve an optimal allocation strategy.

%#####
\paragraph{Fairness Rewards}
In RLHF, rewards are quantified representations of human preferences, reflecting the degree to which model outputs align with human preferences. We model preference learning as a resource allocation problem, where rewards are considered as resources to be allocated. According to Eq (\ref{eq:utility}), our objective is to maximize the utility of reward allocation. We use \( U(\mathbf{a}) \) to measure the extent to which the reward allocation vector \( \mathbf{a} \) aligns with human preferences, with higher values indicating greater utility. Concurrently, it is imperative to ensure fairness in the distribution of rewards. We use $F(\mathbf{a})$ to measure the fairness of the reward allocation vector, with larger values indicating greater fairness. 
%%%%
We expect $F(\mathbf{a})$ to satisfy following properties:
\begin{enumerate}
    \item  \textbf{Continuity}
The fairness measure \( F(\mathbf{a}) \) is continuous on \( \mathbb{R}^n_+ \) for all integers \( n \geq 1 \).

This property ensures that small changes in resource allocation result in only minor changes to the fairness measure, thereby guaranteeing the stability and consistency of the fairness measure.

\item \textbf{Homogeneity}
The fairness measure \( F(\mathbf{a}) \) is a homogeneous function of degree 0:
\[ F(\mathbf{a}) = F(t \cdot \mathbf{a}), \forall t > 0. \]
This property indicates that the fairness measure is independent of the scale of resource allocation.

\item \textbf{Monotonicity}
For \( n = 2 \) entities, the fairness measure \( F(\theta, 1-\theta) \) is monotonically increasing as the absolute difference between the two elements (i.e., \( |1 - 2\theta| \)) shrinks to zero.

This property states that the fairness measure increases as the resource allocation between two entities becomes more equal. 
\end{enumerate}

%%%%

There is a unified fairness metric proposed by \citet{unifair} that satisfies three key properties:
\begin{equation}
f_\tau(\mathbf{a})=\operatorname{sign}(1-\tau) \cdot\left[\sum_{i=1}^n\left(\frac{a_i}{\sum_{j} a_j}\right)^{1-\tau}\right]^{\frac{1}{\tau}},
\label{eq:uni_fair}
\end{equation}
where \(\tau \in \mathbb{R} \) is a constant, which allows for the derivation of different fairness functions based on its value. For instance, 
when $\tau =-1,  f_{\tau=-1}(\mathbf{a})=\frac{(\sum_{i}^{} a_i)^2}{\sum_{i}^{}a_i^2 } =n\cdot J(\mathbf{a} )$ results in Jain's index $J(\mathbf{a})$ \cite{jain1984quantitative}, which is a famous metric for measuring fairness in the resource allocation.

According to Eq (\ref{eq:fair-utility}), we consider the trade-off between utility and fairness. We employ Eq (\ref{eq:uni_fair}) to measure the fairness of reward allocation. Since rewards can be viewed as an infinite resource, and given the property of Homogeneity in fairness metrics, the fairness measure is independent of the unit of measurement or the size of the resource allocation. We can eliminate the constraint on the total amount of resources from Eq (\ref{eq:fair-utility}). Consequently, we propose the following two methods to transform Eq (\ref{eq:fair-utility}) into an unconstrained optimization problem.
\begin{itemize}
    \item \textbf{Fairness Regularization}: We add the two measures together, 
    \begin{equation}
       \max  U(\mathbf{a}) + \alpha F(\mathbf{a}), 
       \label{eq:FR}
    \end{equation}
    where \( \alpha \) is a hyperparameter that controls the impact of the fairness regularization.
    \item \textbf{Fairness Coefficient}: We multiply the two measures, 
    \begin{equation}
       \max U(\mathbf{a}) \cdot F(\mathbf{a})^{\gamma} ,
       \label{eq:FC}
    \end{equation}
  where \( \gamma \) is a hyperparameter that controls the impact of the fairness coefficient.
\end{itemize}

By incorporating these methods, we aim to achieve a trade-off between fairness and utility in the rewards allocation process, ensuring that the rewards not only reflect human preferences accurately but also do so in a fair manner.

\paragraph{Clarification of Fairness}
Finally, we clarify that in the context of LLMs, ``fairness'' usually relates to ``social bias''~\cite{li2023survey,gallegos2024bias}. However, in this paper, we reformulate preference learning from a resource allocation perspective, treating rewards as allocated resources. Here, ``fairness'' refers to the fairness of reward allocation, drawing from resource allocation literature~\cite{kumar2000fairness,unifair}, which differs from social bias. As stated in Section~\ref{sec:intro}, we interpret various reward biases through the lens of reward unfairness, addressing them uniformly. For example, in length bias, ``entity'' refers to data of varying lengths; for category bias, it refers to different data categories, such as ``helpful'' and ``harmless''. In social bias cases, like gender bias, ``entity'' denotes different genders.

\section{Reward-Fairness RLHF}
In this section, we discuss the application of fairness rewards in two scenarios: 
\begin{itemize}
    \item Fairness Rewards for Verification (\S \ref{sec:fair-rm}): We introduce how to train a reward-fairness reward model, which serves as a fair verification. 
    \item Fairness Rewards for RL (\S \ref{sec:fair-rl}): We detail the training of a reward-fairness policy model, which aims to generate outputs that are fairer.
\end{itemize}

\subsection{Fairness Rewards for Verification}\label{sec:fair-rm}

The objective of the reward model is to act as a proxy for human preferences. Typically, it takes a pair of prompt and response \((x, y)\) as input and outputs a scalar score to verify the quality of the pair. To develop a reward-fairness reward model, we need to define the utility function \( U \) and the fairness function \( F \) as per Eq (\ref{eq:FR}) and (\ref{eq:FC}).

For the Bradley-Terry reward model, which uses Eq (\ref{eq:rm_loss}) as its training objective, the goal is to allocate a higher reward to the preferred response \( y_w \) compared to the dispreferred response \( y_l \). Therefore, we define the elements of the allocation vector \( a_i \) as \( a_i = r_{\phi}(y_w) - r_{\phi}(y_l) \). With the allocation vector $\mathbf{a}$ defined, we can directly take Eq (\ref{eq:uni_fair}) as the fairness function, i.e., $F(\mathbf{a})=f_\tau(\mathbf{a})$. The utility function for the reward model is then defined as:
\begin{equation}
U(\mathbf{a}) = \mathbb{E}_{a_i \in \mathbf{a}} \left[ \log \sigma(a_i) \right].
\label{eq:rm_utility}
\end{equation}

We define two types of reward models incorporating fairness: Reward Model with Fairness Regularization (FR RM) and Reward Model with Fairness Coefficient (FC RM). Their training objectives are as follows:

\paragraph{FR RM} The training objective combines the utility and fairness measures additively:
\begin{equation}
\mathcal{L}_{\text{FR RM}} = -\mathbb{E}_{a_i \in \mathbf{a}} \left[ \log \sigma(a_i) \right] - \alpha F(\mathbf{a}).
\label{eq:fr_rm}
\end{equation}

\paragraph{FC RM} The training objective combines the utility and fairness measures multiplicatively:
\begin{equation}
\mathcal{L}_{\text{FC RM}} = -\mathbb{E}_{a_i \in \mathbf{a}} \left[ \log \sigma(a_i) \right] \cdot F(\mathbf{a})^\gamma.
\label{eq:fc_rm}
\end{equation}

\subsection{Fairness Rewards For RL}\label{sec:fair-rl}
Although the training of DPO does not explicitly involve a reward model, it implicitly fits a reward model~\cite{dpo}. We can interpret the term \( \beta \log \frac{\pi_\theta(y \mid x)}{\pi_{\mathrm{ref}}(y \mid x)} \) as an implicit reward. Similar to the reward model, we define the elements of the allocation vector \( a_i \) as:

\begin{equation}
a_i = \beta \log \frac{\pi_\theta(y_w \mid x)}{\pi_{\mathrm{ref}}(y_w \mid x)} - \beta \log \frac{\pi_\theta(y_l \mid x)}{\pi_{\mathrm{ref}}(y_l \mid x)}.
\end{equation}
Consequently, we can derive the same utility function as for the reward model, as shown in Eq (\ref{eq:rm_utility}).
The utility function of the DPO has the same form as that of the reward model, except that the meaning of the allocation vector is slightly different, where \( a_i \in \mathbf{a} \) represents the difference in implicit rewards between the preferred and less preferred responses. With the allocation vector, we can directly use Eq (\ref{eq:uni_fair}) as a fairness function.

We define two types of DPO models incorporating fairness: DPO with Fairness Regularization (FR DPO) and DPO with Fairness Coefficient (FC DPO). The training objectives of $\mathcal{L}_{\text{FR DPO}}$ and $\mathcal{L}_{\text{FC DPO}}$ can be converted to the same form as Eq (\ref{eq:fr_rm}) and (\ref{eq:fc_rm}) respectively.

By incorporating these fairness measures into the DPO framework, we aim to ensure that the model not only aligns with human preferences but also allocate implicit rewards in a fair manner. 

\section{Experiments}
In this section, we empirically investigate the following two research questions $\mathcal{RQ}$:
\begin{itemize}
    \item $\mathcal{RQ}1$: how effective is our Fairness Rewards approach in both verification and RL scenarios? 
    \item $\mathcal{RQ}2$: how does the choice of Fairness Function impact performance? 
\end{itemize}
\begin{figure*}[htbp]
	\centering
	\begin{minipage}[c]{0.32\textwidth}
		\centering
		\includegraphics[width=\textwidth]{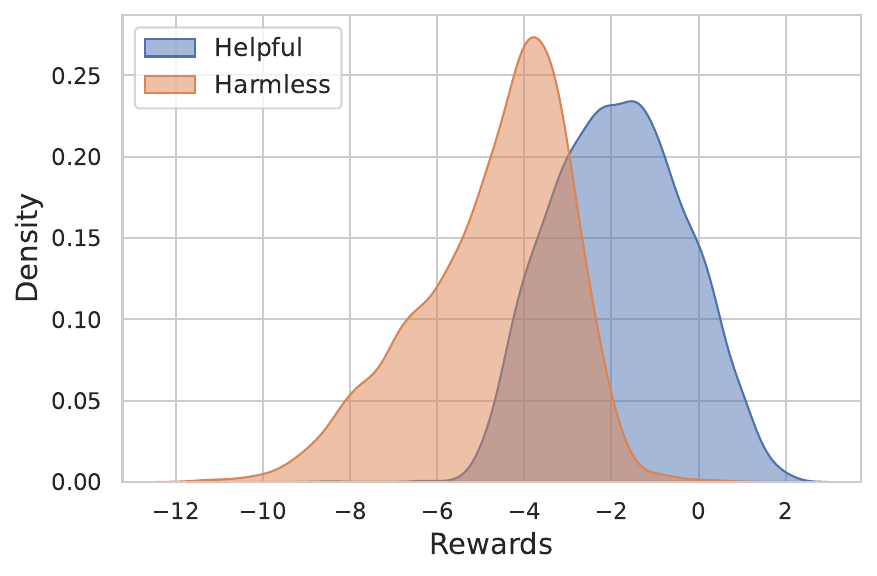}
		\subcaption{BT rewards  on HH-RLHF.}
		% \label{fig_E2_1}
	\end{minipage}
        \begin{minipage}[c]{0.32\textwidth}
    		\centering
    		\includegraphics[width=\textwidth]{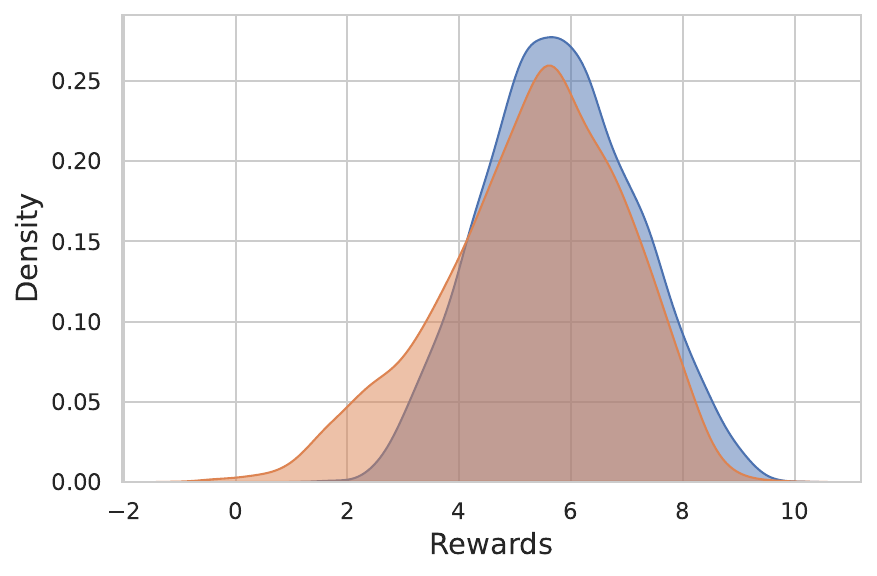}
    		\subcaption{FR rewards  on HH-RLHF.}
    		% \label{fig_E2_1}
    	\end{minipage}
        \begin{minipage}[c]{0.32\textwidth}
    		\centering
    		\includegraphics[width=\textwidth]{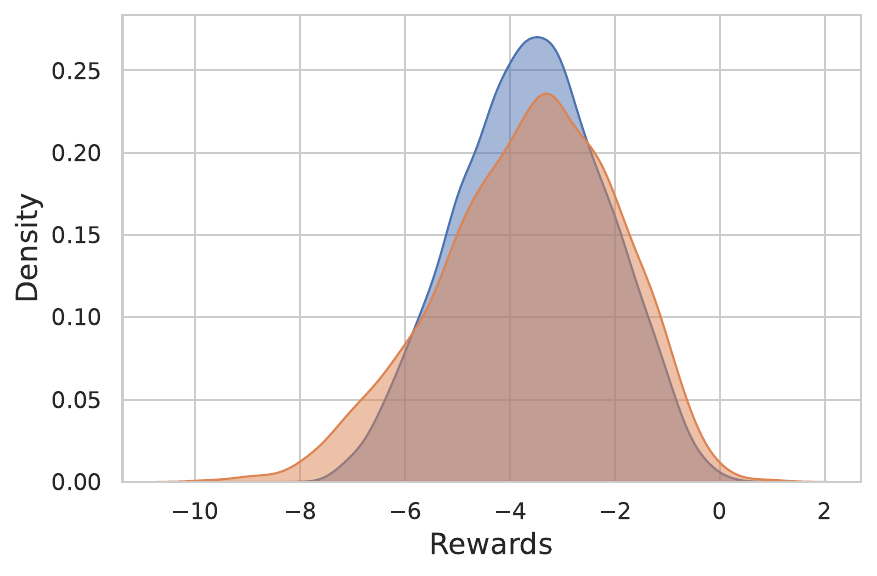}
    		\subcaption{FC rewards  on HH-RLHF.}
    		% \label{fig_E2_1}
    	\end{minipage}
    \\
	\begin{minipage}[c]{0.32\textwidth}
		\centering
		\includegraphics[width=\textwidth]{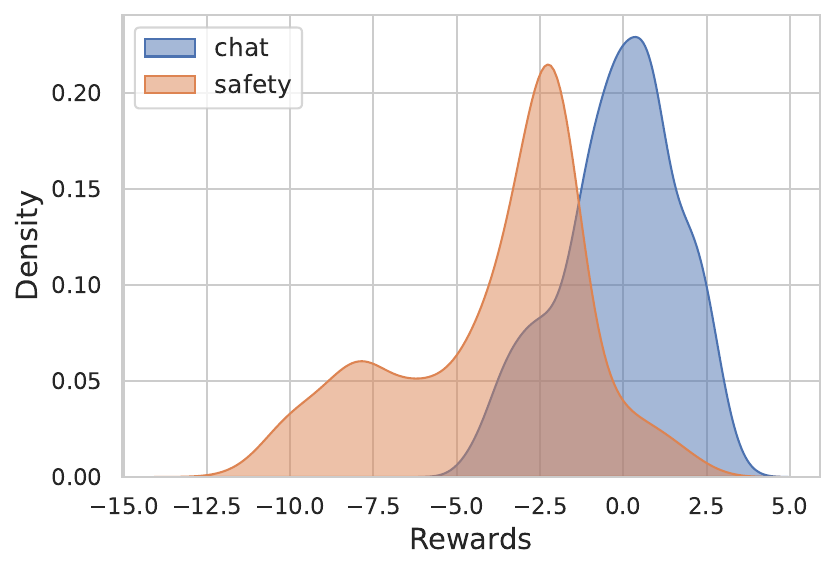}
		\subcaption{BT rewards  on Reward Bench.}
		% \label{fig_E2_2}
	\end{minipage} 
	\begin{minipage}[c]{0.32\textwidth}
		\centering
		\includegraphics[width=\textwidth]{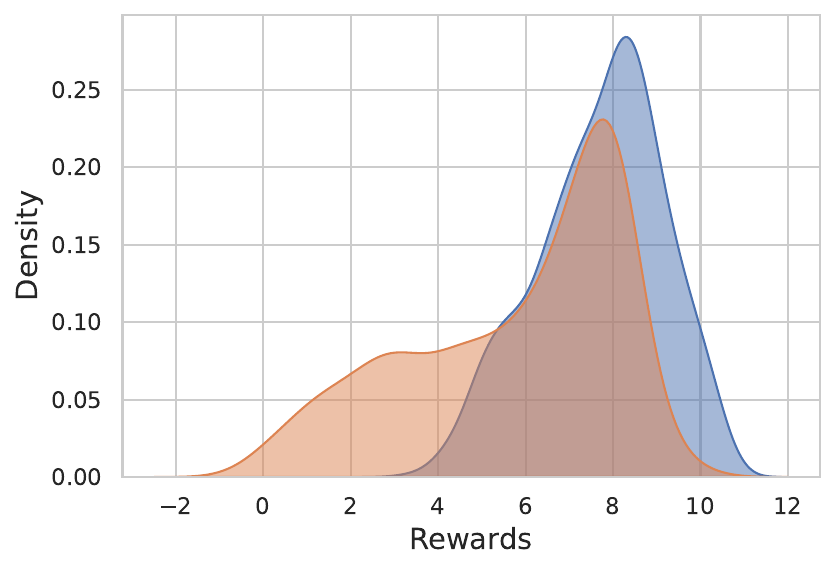}
		\subcaption{FR rewards  on Reward Bench.}
		% \label{fig_E2_3}
	\end{minipage}
        \begin{minipage}[c]{0.32\textwidth}
		\centering
		\includegraphics[width=\textwidth]{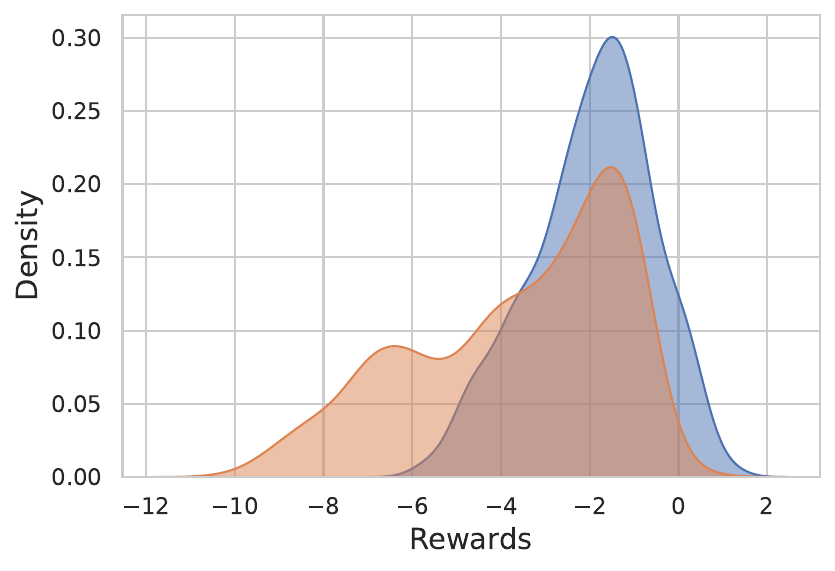}
		\subcaption{FC rewards  on Reward Bench.}
		% \label{fig_E2_3}
	\end{minipage}
	\caption{Rewards  across ID and OOD data.}
	\label{fig:score_Distribution}
\end{figure*}
\paragraph{Datasets \& Baselines} 
In the verification scenario, we conduct experiments on two benchmarks: Reward Bench \cite{rewardbench} and HH-RLHF \cite{hh-rlhf}. Our RMs are trained using the training set from HH-RLHF, making HH-RLHF an in-distribution (ID) benchmark, while Reward Bench serves as an out-of-distribution (OOD) benchmark. We report the accuracy  on both benchmarks. For the RL scenario, we evaluate our methods on the AlpacaEval2 \cite{dubois2024length} and MT-Bench \cite{zheng2023judging} benchmarks. We provide results for AlpacaEval2 in terms of Length-controlled Win Rate (LC WR) and Win Rate (WR), and for MT-Bench, we report the overall score. 
For the policy model, we utilize the UltraFeedback Binarized and SHP datasets for training. We train the policy model using different methods such as DPO \cite{dpo}, KTO \cite{KTO} and R-DPO \cite{disentangling} with HALOs. %, and SimPO \cite{Simpo}. 

% Detailed information about the benchmarks and models can be found in Appendix~\ref{app:setting}. 
\begin{figure}[htb]
    \centering
    \includegraphics[width=0.8\linewidth]{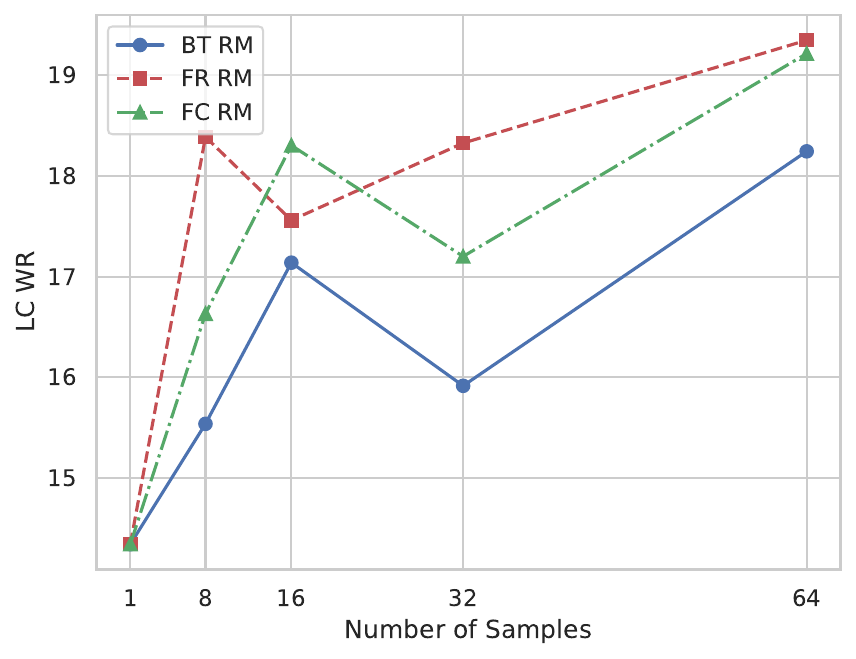}
    \caption{Performance of LLaMA3-SFT on AlpacaEval2 using different verification strategies.}
    \label{fig:sft-bon}
\end{figure}
\paragraph{Implementation Details} 
For the reward model, we train on the HH-RLHF training set for one epoch with a learning rate of \(2 \times 10^{-6}\). 
For the policy model, we utilize the UltraFeedback Binarized and SHP datasets, training for one epoch with a learning rate of \(5 \times 10^{-6}\). During sampling with the policy model, the temperature coefficient is set to 1. All experiments are performed on an 8 $\times$ H800 machine. 
Both the reward models and the policy models are trained using LLaMA3-SFT (a base model developed by \citet{rlhfflow}) and Qwen2.5-SFT (a base model we trained following \citet{rlhfflow}). 
Further experimental details can be found in Appendix~\ref{app:setting}.

\subsection{Main Results ($\mathcal{RQ}1$)}
\subsubsection{Fairness Verification}
\begin{figure*}[htbp]
	\centering
	\begin{minipage}[c]{0.32\textwidth}
		\centering
		\includegraphics[width=\textwidth]{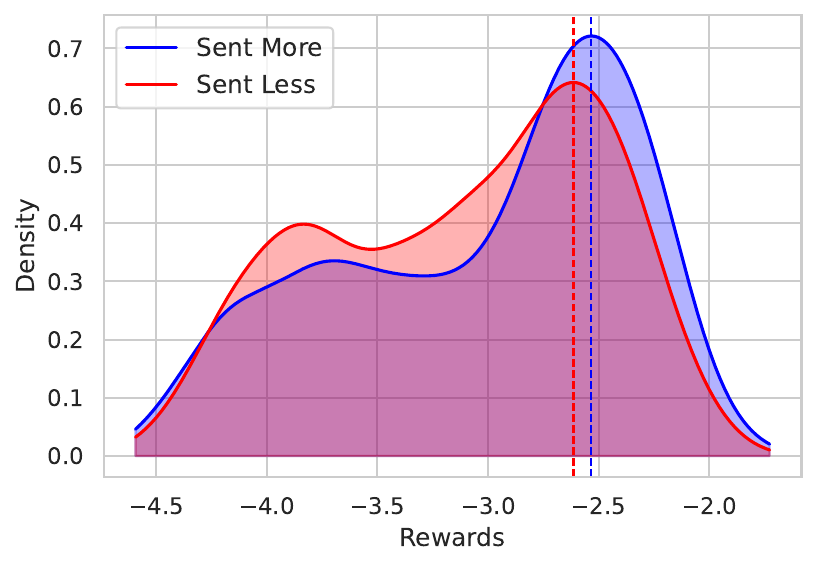}
		\subcaption{Rewards distribution of BT RM.}
		% \label{fig_E2_1}
	\end{minipage}
        \begin{minipage}[c]{0.32\textwidth}
    		\centering
    		\includegraphics[width=\textwidth]{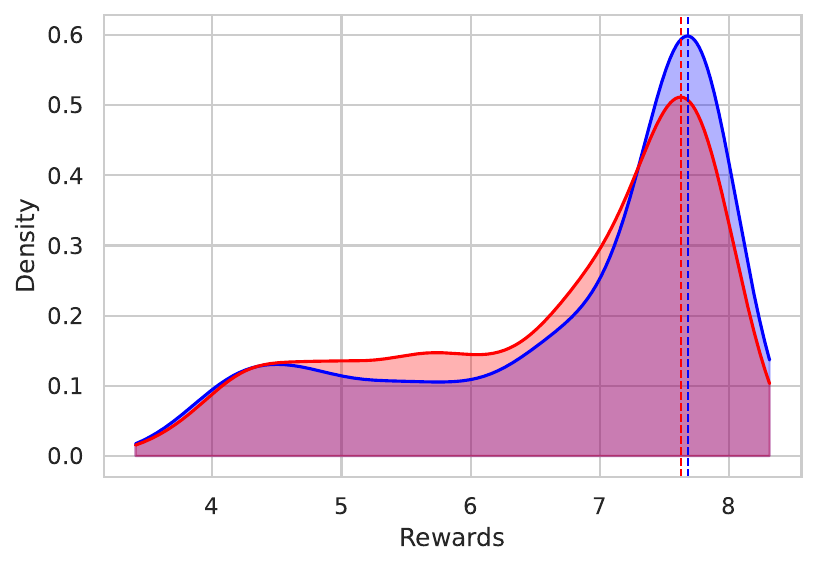}
    		\subcaption{Rewards distribution of FR RM.}
    		% \label{fig_E2_1}
    	\end{minipage}
        \begin{minipage}[c]{0.32\textwidth}
    		\centering
    		\includegraphics[width=\textwidth]{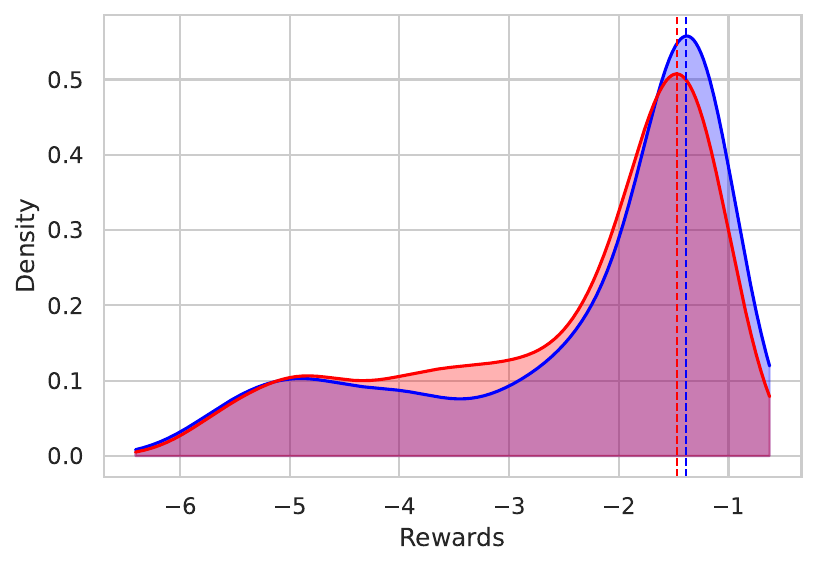}
    		\subcaption{Rewards distribution of FC RM.}
    		% \label{fig_E2_1}
    	\end{minipage}
	\caption{Rewards distribution on CrowS-Pairs.}
	\label{fig:soical_bias}
\end{figure*}

\begin{table*}[htb]
\centering
\caption{Performance of different verification strategies on two reward model benchmarks.}
\label{tab:rm_res}
\begin{tabular}{@{}l|ccccc|ccc@{}}
\toprule
\multirow{2}{*}{Verifiers} & \multicolumn{5}{c|}{Reward Bench}               & \multicolumn{3}{c}{HH-RLHF} \\ \cmidrule(l){2-9} 
                           & Chat   & Chathard & Reasoning & Safety & Avg.   & Helpful  & Harmless & Avg.  \\ \midrule
BT RM                        & 93.02  & 57.02    & 84.98     & 77.43  & 78.11  & 74.38    & 73.23    & 73.81 \\
FR RM                   & 94.41  & 57.02    & 83.86     & 78.24  & 78.38  & 73.49    & 73.62    & 73.55 \\
FC RM                   & 94.41  & 53.29    & 85.53     & 76.76  & 77.50  & 74.30    & 73.62    & 73.96 \\ \bottomrule
\end{tabular}
\end{table*}

Figure~\ref{fig:score_Distribution} illustrates the distribution of rewards from different RMs on ID data (HH-RLHF) and OOD data (Reward Bench). The first row of figures shows the rewards distribution on the ID dataset HH-RLHF. It is evident that the Bradley-Terry (BT) RM exhibits a significant disparity in the distribution of rewards between Helpful and Harmless data, indicating an unfair allocation of rewards. In contrast, Reward Model with Fairness Regularization (FR RM) and Reward Model with Fairness Coefficient (FC RM) demonstrate a more consistent rewards distribution across Helpful and Harmless data, indicating that those two RMs are fairer. The figures in the second row show the distribution of rewards on the OOD data, and we can draw the same conclusions as for the ID data.
Table~\ref{tab:rm_res} presents the performance of the three RMs on the Reward Bench and HH-RLHF. The results show no significant performance difference between FR RM, FC RM, and BT RM, suggesting that Fair RMs achieve a good trade-off between fairness and utility without sacrificing model performance. Additionally,we provide the distribution of rewards on length in Appendix \ref{app:exp}.

\paragraph{Data Selection} We will show that this fair distribution of rewards will bring extra benefits in data selection. Figure~\ref{fig:sft-bon} shows the performance on AlpacaEval2 when using different RMs to select samples. From the figure, we can draw two conclusions: (1) When sampling the same number of samples, FR RM and FC RM can select higher quality samples compared to BT RM. (2) To achieve the same performance, FR RM and FC RM require fewer samples, indicating higher sampling efficiency.

\paragraph{Social Bias}

We further validated our methods on the CrowS-Pairs\footnote{\url{https://github.com/nyu-mll/crows-pairs/}} \cite{nangia2020crows} dataset, which includes sentences with social biases. This dataset contains two types of sentences: ``sent more", which is more stereotypical, and ``sent less", which is less stereotypical. It encompasses nine types of biases, such as gender and nationality. As shown in Figure \ref{fig:soical_bias}, the BT RM tends to assign higher rewards to the more stereotypical sentences, resulting in a larger distributional difference between ``sent more" and ``sent less". This indicates that the BT RM exhibits unfairness across various social biases. In contrast,  FR RM and FC RM show smaller distributional differences, demonstrating greater fairness across different social biases.

These findings highlight the effectiveness of our Fair RMs in providing a fairer reward distribution while maintaining high performance and sampling efficiency. %Further experimental details and analysis can be found in Appendix~\ref{app:exp}.

\subsubsection{Fairness Policy Model}

% \usepackage{booktabs}
% \usepackage{multirow}
\begin{table}[htb]
\centering
\caption{Performance of different policy models on AlpacaEval2 and MT-Bench.}
\label{tab:dpo_res}
\resizebox{.49\textwidth}{!}{
\begin{tabular}{@{}ll|cc|c@{}}
\toprule
 &       & \multicolumn{2}{c|}{AlpacaEval2} & MT-Bench      \\ \cmidrule(l){3-5} 
 &       & LC WR           & WR             & Overall \\ \midrule
 & SFT   & 14.34           & 8.17           & 5.93          \\
 & R-DPO & {\ul 20.87}     & 11.16          & 6.48          \\
 & KTO   & 19.44           & {\ul 16.64}    & {\ul 6.64}    \\ \cmidrule(l){2-5}
 & DPO   & 16.71           & 14.23          & 6.46          \\  
 &
  \multicolumn{1}{r|}{\cellcolor[HTML]{C0C0C0}+FR} &
  \cellcolor[HTML]{C0C0C0}20.48 &
  \cellcolor[HTML]{C0C0C0}15.74 &
  \cellcolor[HTML]{C0C0C0}\textbf{6.70} \\
\multirow{-6}{*}{LLaMA3} &
  \multicolumn{1}{r|}{\cellcolor[HTML]{C0C0C0}+FC} &
  \cellcolor[HTML]{C0C0C0}\textbf{21.10} &
  \cellcolor[HTML]{C0C0C0}\textbf{16.96} &
  \cellcolor[HTML]{C0C0C0}6.58 \\ \midrule
 & SFT   & 13.47           & 8.11           & 5.69          \\
 & R-DPO & {\ul 19.95}     & 10.15          & {\ul 7.05}    \\
 & KTO   & 17.81           & 14.39          & 6.72          \\ \cmidrule(l){2-5} 
 & DPO   & 18.93           & 13.18          & 6.59          \\ 
 &
  \multicolumn{1}{r|}{\cellcolor[HTML]{C0C0C0}+FR} &
  \cellcolor[HTML]{C0C0C0}\textbf{21.05} &
  \cellcolor[HTML]{C0C0C0}\textbf{15.25} &
  \cellcolor[HTML]{C0C0C0}\textbf{7.24} \\
\multirow{-6}{*}{Qwen2.5} &
  \multicolumn{1}{r|}{\cellcolor[HTML]{C0C0C0}+FC} &
  \cellcolor[HTML]{C0C0C0}19.72 &
  \cellcolor[HTML]{C0C0C0}{\ul 14.53} &
  \cellcolor[HTML]{C0C0C0}7.00 \\ \bottomrule
\end{tabular}
}
\end{table}

Table~\ref{tab:dpo_res} presents the results of different policy models on AlpacaEval2 and MT-Bench. It can be observed that our fairness reward methods, when combined with DPO, consistently demonstrates superior performance on AlpacaEval2 and MT-Bench with both LLaMA3-SFT and Qwen2.5-SFT as base models. This highlights the effectiveness of our fairness rewards method. The success of our methods can be attributed to their ability to implicitly fit a fairness RM during the training of policy models, thereby generating higher quality outputs. 

\begin{figure}[htb]
    \centering
    \includegraphics[width=0.8\linewidth]{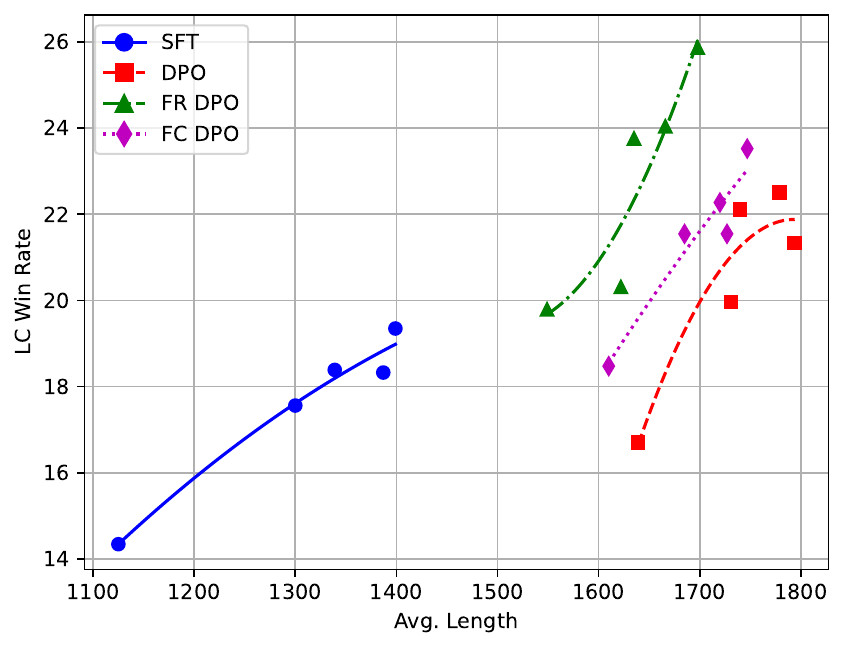}
    \caption{Length and performance relationships of samples for different models.}
    \label{fig:samples_length_bias}
\end{figure}

Combining Fair DPOs with Fair RM further enhances performance. We sample the policy model 1, 8, 16, 32, and 64 times, using Fair RM to select the best sample. We recorded the lengths and performance of these samples and fitted a curve to this data, as shown in Figure \ref{fig:samples_length_bias}. It can be observed that for the same model, performance gradually increases with length, indicating a correlation between performance and length. However, for different models, aligned models produce higher quality outputs compared to the SFT model, but their outputs are also longer. Among the three aligned models, Fair DPOs achieve better performance than DPO while producing shorter outputs, suggesting that our model can mitigate length bias to some extent.

\subsection{Ablation Study ($\mathcal{RQ}2$)}
\paragraph{Impact of Fairness Function}
% \usepackage{booktabs}
% \usepackage{multirow}
\begin{table}[htb]
\centering
\caption{Performance under different fairness functions.}
\label{tab:ablation_tau}
\begin{tabular}{@{}l|cc|c@{}}
\toprule
\multirow{2}{*}{Model} & \multicolumn{2}{c|}{AlpacaEval2} & MT-Bench      \\ \cmidrule(l){2-4} 
                       & LC WR           & WR             & Overall Score \\ \midrule
DPO                    & 16.71           & 14.23          & 6.46          \\ \midrule
FR DPO                 &                 &                &               \\
$\tau=-5$                 & 19.72           & 14.44          & 6.59          \\
$\tau=-1 $                & 20.48           & 15.74          & 6.70          \\
$\tau=0.5  $              & 20.01           & 15.21          & 6.56          \\
$\tau=2  $                & 20.01           & 17.35          & 6.69          \\
$\tau=10 $                & 19.98           & 16.24          & 6.62          \\ \bottomrule
\end{tabular}
\end{table}

Eq (\ref{eq:uni_fair}) presents a unified metric for measuring fairness, from which different fairness functions can be derived by varying $\tau$. We aim to explore the impact of different fairness functions on performance. We experiment with various $\tau$ values within the range of [-5, 10], and the results are summarized in Table \ref{tab:ablation_tau}. It can be observed that the performance of FR DPO consistently surpasses that of the native DPO across all fairness functions. This indicates that our method is robust to variations in $\tau$. The reason for this robustness is that all fairness functions derived from the unified metric satisfy the three desired properties, ensuring that the rewards obtained are fair.

\paragraph{Impact of Fairness Contribution $\alpha$}
\begin{figure}[htb]
    \centering
    \includegraphics[width=0.85\linewidth]{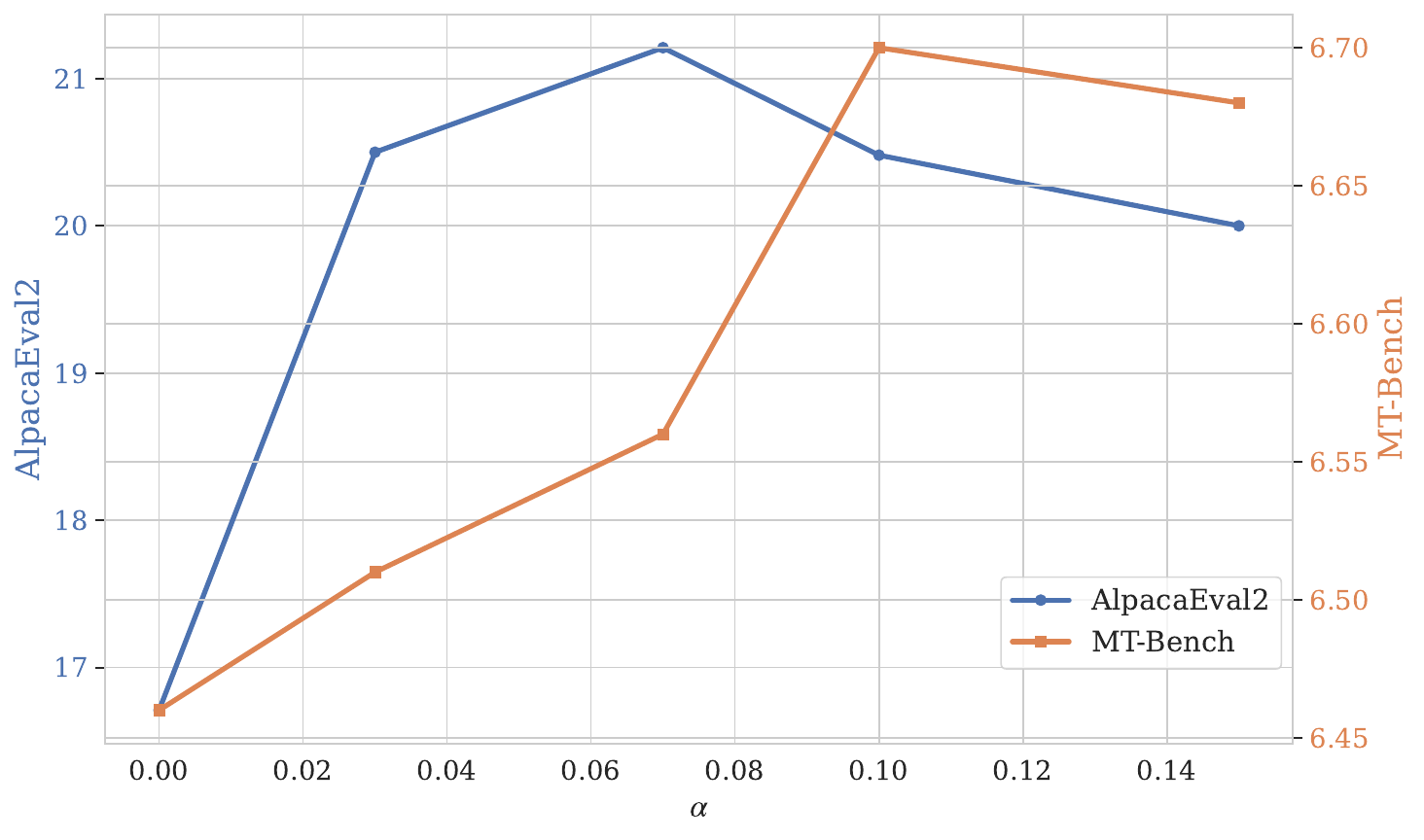}
    \caption{Performance under different fairness contribution $\alpha$.}
    \label{fig:ablation_alpha}
\end{figure}
We conduct experiments by fixing \(\tau = -1\) and varying the fairness contribution \(\alpha\) within the range of [0, 0.15], as illustrated in Figure \ref{fig:ablation_alpha}. When \(\alpha = 0\), the model reduces to the native DPO. As \(\alpha\) increases, the model's performance on AlpacaEval2 and MT-Bench initially improves and then declines. This trend occurs because, at lower values of \(\alpha\), enhancing fairness contributes to better output quality. However, when \(\alpha\) becomes too large, the excessive emphasis on fairness leads to a compromise in utility. Considering the trade-off between fairness and utility, we typically set \(\alpha = 0.1\) in practical experiments. We present the ablation experiments on the Fairness Contribution $\gamma$ in Appendix~\ref{app:exp}.

\section{Conclusion}

In this paper, we tackle the critical issue of reward unfairness in RLHF. We identify that length bias and category bias are specific case of the broader problem of reward unfairness. To address this comprehensively, we introduce the reward fairness framework, which models preference learning as a resource allocation problem to balance fairness and utility in reward distribution.
We propose two methods to achieve fairness rewards: Fairness Regularization and Fairness Coefficient. These methods are applied in two key scenarios: training a fairness RM for verification and training a policy model for reinforcement learning that implicitly incorporates fair rewards.

\section*{Limitation}
We investigate the issue of rewards\ unfairness and proposed a solution from the perspective of resource allocation, validating our approach in both verification and RL scenarios. The limitations of this study are summarized as follows: (1) Reward unfairness is a broad concept, and this paper primarily focuses on category bias and length bias, with a simple validation on social bias. However, reward unfairness may be related to various issues in reward models, such as reward hacking. (2) Our Fairness Rewards method can seamlessly integrate with RM and RL frameworks that utilize RMs either explicitly or implicitly. We have only validated it on BT models and DPO, but the Fairness Rewards method has the potential for broader applications.

\section*{Acknowledgements}
We would like to express our sincere gratitude to all the anonymous reviewers for their invaluable feedback that greatly improved this paper. In particular, special thanks to Dr. Weiran Shen for his insightful suggestions and invaluable assistance in early-stage discussions.
This research was supported by National Natural Science Foundation of China (No.62476277), National Key Research and Development Program of China (NO. 2024YFE0203200), CCF-ALIMAMA TECH Kangaroo Fund(No.CCF-ALIMAMA OF 2024008), and Huawei-Renmin University joint program on Information Retrieval. We also acknowledge the support provided by the fund for building worldclass universities (disciplines) of Renmin University of China and by the funds from Beijing Key Laboratory of Big Data Management and Analysis Methods, Gaoling School of Artificial Intelligence, Renmin University of China, from Engineering Research Center of Next-Generation Intelligent Search and Recommendation, Ministry of Education, from Intelligent Social Governance Interdisciplinary Platform, Major Innovation \& Planning Interdisciplinary Platform for the “DoubleFirst Class” Initiative, Renmin University of China, from Public Policy and Decision-making Research Lab of Renmin University of China, and from Public Computing Cloud, Renmin University of China.

\bibliography{custom}

\appendix
\onecolumn
\section{Further Discussions}

\paragraph{Further explanation of Figure~\ref{fig:introduction}}
Someone may argue that `` \( y_1^l \) could be a better response than \( y_2^w \)''.
Our objective is to achieve both utility and fairness in reward allocation in RLHF. Figure~\ref{fig:overview} in our paper effectively illustrates our objective. Considering an  example where four responses ($y_1, y_2, y_3, $ and $y_4$) are sampled for the same prompt, with rewards ranked as $y_1 > y_2 > y_3 > y_4$. For preference data pairs ($y_1 , y_2$) and ($y_3, y_4$), it is reasonable that $y_2 > y_3$. This does not conflict with our objective, as utility measures the alignment of reward allocation with human preferences, meaning that a good response should receive a higher reward than a poor one. Fairness focuses on the reward distribution across different types of samples, typically from various domains. These absolute reward values are generally incomparable, but we expect their distributions to be as fair as possible to avoid issues in downstream scenarios such as rejected sampling and RL. Unfortunately, our experiments reveal that the commonly used BT model exhibits pervasive reward unfairness, as shown in Figure~\ref{fig:score_Distribution}  and  Figure~\ref{fig:length_bias_app}.  This unfairness affects both in-distribution and out-of-distribution data, leading to category and length biases.
Additional, Table~\ref{fig:avg_reward_app} presents the average rewards from the Bradley-Terry (BT) model on the HH-RLHF dataset. For both ``helpful'' and ``harmless'' data, the rewards for ``chosen'' are greater than those for ``rejected'', aligning with the utility objective. However, the rewards for ``helpful'' are significantly higher than those for ``harmless'', which is unfair. When such unfair rewards are used in rejected sampling and RL, the model's output becomes more helpful but neglects harmlessness.

\begin{table}[htb]
\centering
\caption{Average rewards of BT model on the HH-RLHF dataset.}
\label{fig:avg_reward_app}
\begin{tabular}{@{}l|cc|cc@{}}
\toprule
            & \multicolumn{2}{c|}{Helpful} & \multicolumn{2}{c}{Harmless} \\ \cmidrule(l){2-5} 
            & chosen       & rejected      & chosen       & rejected      \\ \midrule
Avg. Reward & -1.39        & -2.26         & -4.15        & -5.23         \\ \bottomrule
\end{tabular}
\end{table}

\paragraph{Fairness and Utility}
Figure~\ref{fig:overview} illustrates our dual objectives: achieving fairness and utility in reward allocation. The relationship between these objectives varies slightly between verification and reinforcement learning scenarios. In the verification scenario, ``fairness'' aims to make the distribution of different types of rewards more consistent, while ``utility'' ensures that for any given prompt, the reward for a good response exceeds that for a bad response. These objectives are inherently independent and non-conflicting, though balancing them necessitates a multi-objective optimization. In the reinforcement learning scenario, "fairness" can even enhance "utility" by guiding the model's output to be both more helpful and harmless.

\section{Experiment Setting}
\label{app:setting}

\paragraph{Datasets \& Baselines} 

In the verification scenario, we conduct experiments on two benchmarks: Reward Bench \cite{rewardbench} and HH-RLHF \cite{hh-rlhf}. Our RMs are trained using the training set from HH-RLHF\footnote{\url{https://huggingface.co/datasets/Anthropic/hh-rlhf}}, making HH-RLHF an in-distribution (ID) benchmark, while Reward Bench serves as an out-of-distribution (OOD) benchmark. We report the accuracy  on both benchmarks. For the RL scenario, we evaluate our methods on the AlpacaEval2 \cite{dubois2024length} and MT-Bench \cite{zheng2023judging} benchmarks. We provide results for AlpacaEval2 in terms of Length-controlled Win Rate (LC WR) and Win Rate (WR), and for MT-Bench, we report the overall score. For the policy model, we utilize the UltraFeedback Binarized\footnote{\url{https://huggingface.co/datasets/HuggingFaceH4/ultrafeedback_binarized}} and SHP\footnote{\url{https://huggingface.co/datasets/stanfordnlp/SHP}} datasets for training. We train the policy model using different methods such as DPO \cite{dpo}, KTO \cite{KTO} and R-DPO \cite{disentangling} with HALOs\footnote{\url{https://github.com/ContextualAI/HALOs/}}. %, and SimPO \cite{Simpo}. 

\paragraph{Training Setting}
For the reward model, we train on the HH-RLHF training set for one epoch with a learning rate of \(2 \times 10^{-6}\) and a global batch size of 256. 
For the policy model, we utilize the UltraFeedback Binarized and SHP datasets, training for one epoch with a learning rate of \(5 \times 10^{-6}\) and a global batch size of 256. During sampling with the policy model, the temperature coefficient is set to 1. All experiments are performed on an 8 $\times$ H800 machine. 
Both the reward models and the policy models are trained using LLaMA3-SFT\footnote{\url{https://huggingface.co/RLHFlow/LLaMA3-SFT}} (a base model developed by \citet{rlhfflow}) and Qwen2.5-SFT (a base model we trained\footnote{\url{https://github.com/RLHFlow/Online-RLHF/}} following \citet{rlhfflow}). 
Qwen2.5-SFT is a model we trained based on the Qwen2.5-7B base using the dataset from RLHFow\footnote{\url{https://huggingface.co/datasets/RLHFlow/RLHFlow-SFT-Dataset-ver2}} for one epoch. The 
global batch size was set to 128, and the learning rate was \(2 \times 10^{-5}\). For all policy models, the \(\beta\) parameter was uniformly set to 0.1. Additionally, the desirable weight and undesirable weight for the KTO were both set to 1.

\section{Supplement Experiment}\label{app:exp}

\begin{figure}[htb]
    \centering
    \begin{minipage}[c]{0.32\textwidth}
		\centering
		\includegraphics[width=\textwidth]{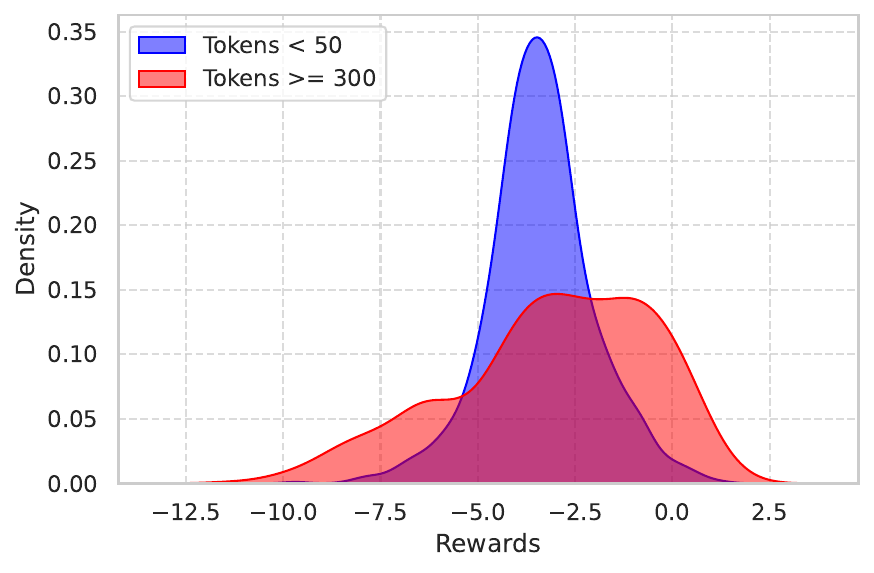}
		\subcaption{BT rewards on length.}
		% \label{fig:length_bias}
	\end{minipage}
    \begin{minipage}[c]{0.32\textwidth}
		\centering
		\includegraphics[width=\textwidth]{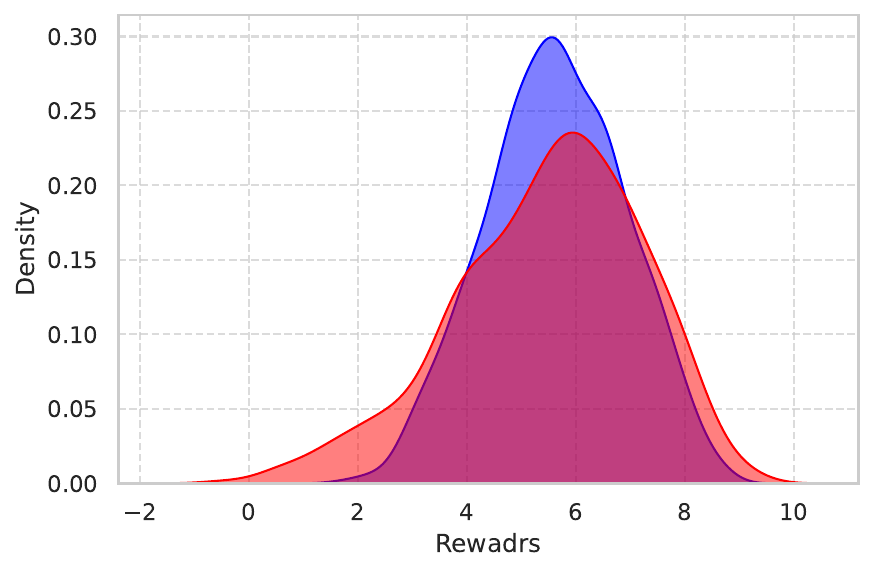}
		\subcaption{FR rewards on length.}
		% \label{fig:category_bias}
	\end{minipage}
    \begin{minipage}[c]{0.32\textwidth}
		\centering
		\includegraphics[width=\textwidth]{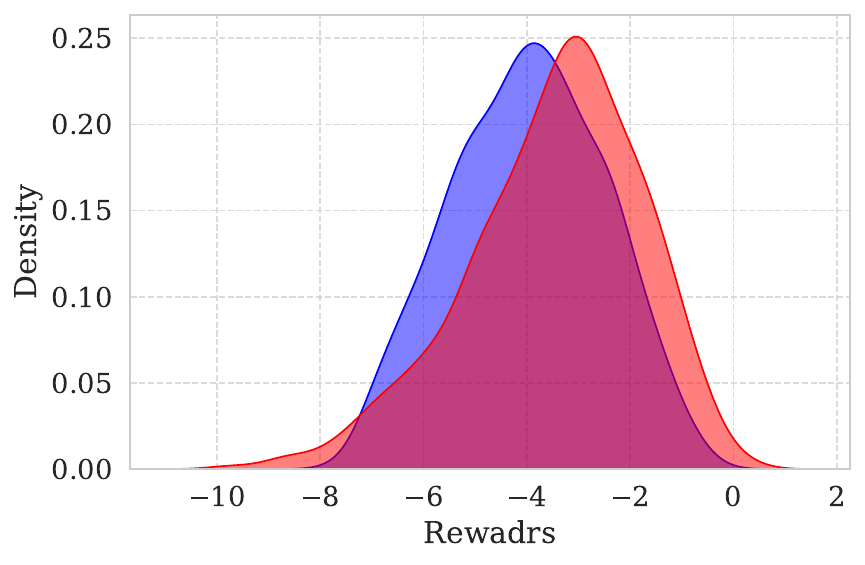}
		\subcaption{FC rewards on length.}
		% \label{fig:category_bias}
	\end{minipage}
    \caption{Rewards on Length}
    \label{fig:length_bias_app}
\end{figure}

\paragraph{Rewards on Length} The rewards distribution of BT RM, FR RM, and FC RM across different lengths on the HH-RLHF dataset is illustrated in Figure \ref{fig:length_bias_app}. Our FR RM and FC RM exhibits a more consistent rewards distribution across varying lengths, demonstrating that our method effectively mitigates length bias.

\begin{figure}[htb]
    \centering
    \includegraphics[width=0.65\linewidth]{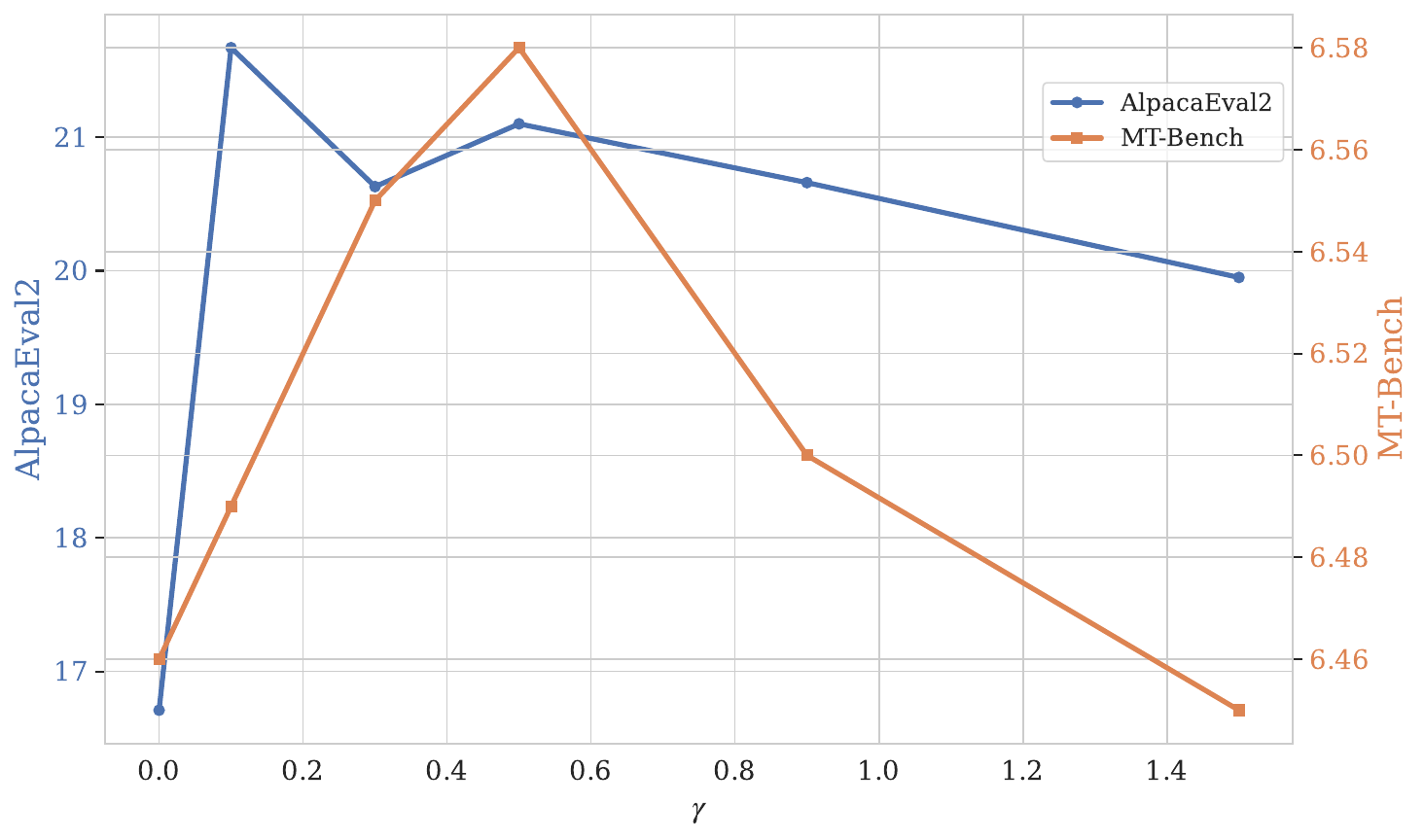}
    \caption{Performance under different fairness contribution $\gamma$.}
    \label{fig:ablation_gamma}
\end{figure}

\paragraph{Impact of Fairness Contribution $\gamma$}
We conduct experiments by fixing \(\tau = -1\) and varying the fairness contribution \(\gamma\) within the range of [0, 1.5], as illustrated in Figure \ref{fig:ablation_gamma}. We obtain conclusions similar to those with fairness contribution \(\alpha\). Considering the trade-off between fairness and utility, we typically set \(\gamma = 0.5\) in practical experiments.

\end{document}